\documentclass[letterpaper, 10 pt, conference]{ieeeconf}

\IEEEoverridecommandlockouts
\usepackage{amsmath}
\usepackage{amssymb}
\usepackage{amsfonts}
\usepackage{bm}
\usepackage{graphicx}
\usepackage{cite}
\usepackage{booktabs}
\usepackage{multirow}
\usepackage{array}
\usepackage{tikz}
\usepackage[dvipsnames]{xcolor}
\usepackage{url}
\usepackage[table]{xcolor}    
\usepackage{float}            
\usepackage{stfloats}         
\usepackage{etoolbox}
\usepackage{hyperref}

\usepackage{algorithm}
\usepackage{algpseudocode}
\usetikzlibrary{calc}

\hypersetup{
    colorlinks=true,       
    linkcolor=blue,        
    filecolor=magenta,     
    urlcolor=cyan,         
    pdftitle={Your Title}  
}

\makeatletter
\patchcmd{\@makecaption}
    {\\}
    {:\nobreakspace}
    {}
    {\PackageWarning{IEEEcaption}{Table caption patch failed}}
\makeatother

\title{\LARGE \bf Pareto-Optimal Entropy-Regularized Trajectory Optimization}

\author{Dimitrios S. Georgiou$^{1}$, Augustinos D. Saravanos$^{2}$, and Evangelos A.  Theodorou$^{1}$%
\thanks{$^{1}$Daniel Guggenheim School of Aerospace Engineering, Georgia Institute of Technology, Atlanta, GA, USA. {\tt\footnotesize\{dgeorgiou3, evangelos.theodorou\}@gatech.edu}}%
\thanks{$^{2}$Department of Aeronautics and Astronautics, Massachusetts Institute of Technology, Cambridge, MA, USA. {\tt\footnotesize asaravan@mit.edu}}%
\thanks{Corresponding author: Dimitrios Georgiou}}

\begin{document}

\maketitle
\thispagestyle{empty}
\pagestyle{empty}

\begin{abstract}
Trajectory optimization (TO) under nonlinear dynamics, actuation limits and collision avoidance constraints is a fundamental problem in robotics, albeit especially challenging due to its highly non-convex nature. 
For this setting, Differential Dynamic Programming (DDP) is an efficient second-order shooting method, yet its local structure renders it vulnerable to suboptimal basins. 
Sampling-augmented variants mitigate this susceptibility through stochastic exploration, but often sample only around the few trajectories they retain for reoptimization, based solely on their cost which restricts exploration breadth. 
We introduce \underline{P}areto-Optimal \underline{E}ntropy-\underline{R}egularized DDP (PER-DDP), an entropy-regularized population framework derived from the free-energy/relative-entropy inequality.
Our method combines prior-guided sampling that shapes exploration around each retained trajectory, with expanded rollout evaluations that probe these sampling policies beyond the few retained candidates, and Pareto filtering for preserving task--constraint alternatives across iterations. 
This decouples sampling effort from the optimization population size and broadens exploration without sacrificing the second-order structure that makes DDP effective. 
Across multiple systems and hundreds of environments, PER-DDP achieves higher success rates than state-of-the-art sampling-augmented TO methods and finds reliable solutions in environments beyond the reach of all baselines. A supplemental video can be found \href{https://youtu.be/_ICx3OK3x60}{here}.

\vspace{-1mm}
\end{abstract}



\section{Introduction}

Autonomous robots must achieve task goals while respecting nonlinear dynamics, actuation limits, and safety constraints, a problem commonly cast as trajectory optimization (TO).
%
However, the induced optimization landscape is often highly non-convex: in cluttered environments, for example, distinct local basins correspond to qualitatively different motion strategies.
Consequently, local optimization methods commit to one such strategy early, \textit{e.g.}, to one side of an obstacle, and cannot recover the alternatives no matter how well they converge.
Other popular approaches, such as sampling-based motion planning \cite{kavraki1996probabilistic} and reinforcement learning~\cite{kober2013reinforcement,sutton2018reinforcement} sidestep this issue through global search and offline training, respectively.
TO instead exploits available models and constraints directly, producing locally optimal, dynamically feasible trajectories, at the price of confronting the non-convex landscape within the optimization itself.

Derivative-based TO methods such as Sequential Quadratic Programming (SQP) \cite{nocedal2006numerical} and Differential Dynamic Programming (DDP) \cite{JacobsonMayne1970DDP} exploit gradients and curvature to progress rapidly toward local solutions. 
DDP is an especially appealing and scalable algorithm as it exploits the sequential structure of the dynamics through local second-order models, yielding control updates and feedback gains with linear complexity in the horizon.
These local models, however, are constructed around a single nominal trajectory, rendering DDP prone to suboptimal local minima \cite{saravanos2023distributed}.
Escaping such suboptimal basins thus requires exploring alternative trajectories without sacrificing the local second-order structure that makes DDP effective.

 \begin{figure}[t]
        \centering
        \setlength{\abovecaptionskip}{-1pt}
    
        \begin{tikzpicture}
            \node[anchor=south west, inner sep=0] (img) at (0,0)
            {\includegraphics[width=0.49\textwidth]{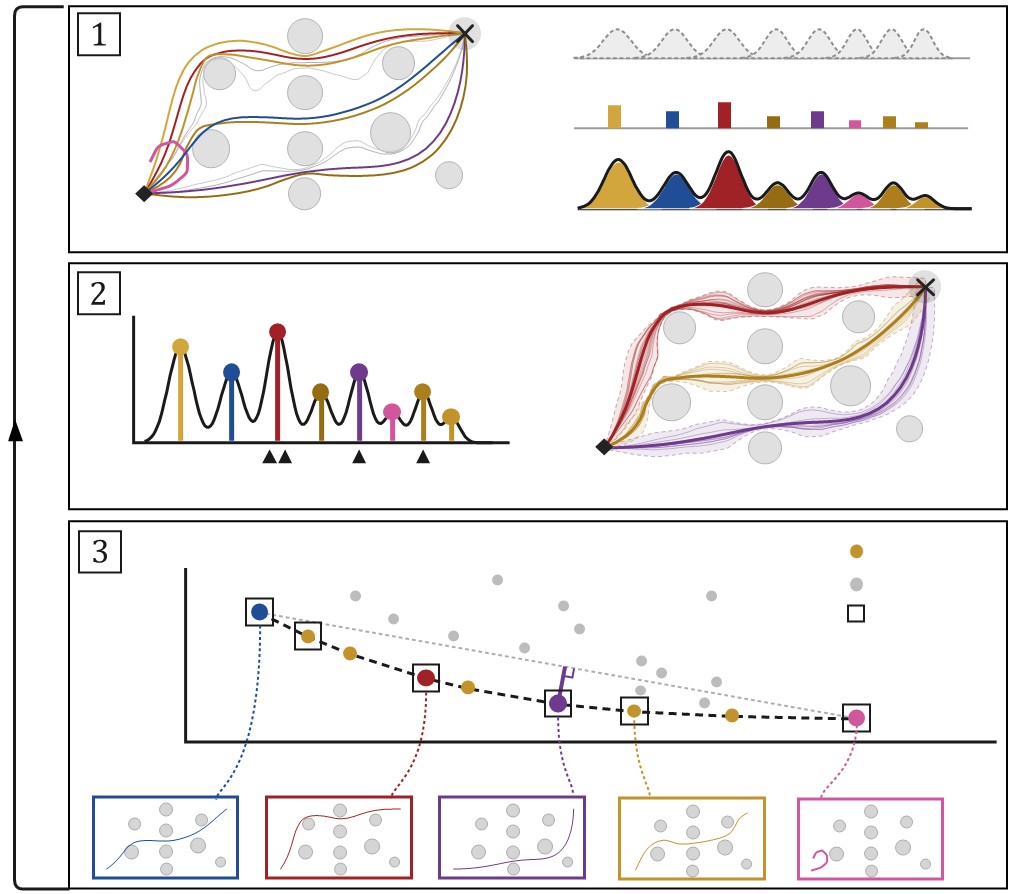}};
    
            \begin{scope}[
                x={(img.south east)},
                y={(img.north west)},
                every node/.style={font=\footnotesize}
            ]
    
    
            \node[font=\scriptsize] at (0.29,0.740)
            {\textbf{\textit{Optimize}} \(N\) trajectories with DDP};
    
            \node[font=\scriptsize] at (0.745,0.740)
            {\textbf{Prior-guided} Sampling Policy};
    
            \node[anchor=west] at (0.510,0.950)
            {\(p^{(r)}\)};
    
            \node[anchor=west] at (0.510,0.880)
            {\(w^{(r)}\)};
    
            \node[anchor=west] at (0.520,0.780)
            {\(q^\star\)};
    
            \node[font=\normalsize] at (0.936,0.890) {\(\times\)};
            \node[font=\normalsize] at (0.936,0.810) {\(=\)};
    
            \node[anchor=west] at (0.125,0.661)
            {\(q^\star\)};
    
            \node[font=\scriptsize] at (0.286,0.449)
            {\textbf{Categorical} Distribution};
    
            \node[font=\scriptsize] at (0.740,0.449)
            {\textbf{\textit{Sample}} \(M\) rollouts around \(R\) modes};
    
            \node[font=\scriptsize, anchor=west] at (0.120,0.380)
            {Pareto Front \(\mathcal{F}_{P}\)};
    
            \node[rotate=90, font=\normalsize] at (0.145,0.255)
            {\(J_g\)};
    
            \node[font=\normalsize] at (0.952,0.140)
            {\(J\)};
    
            \node[font=\scriptsize, anchor=west] at (0.835,0.377) {Fillers};
            \node[font=\scriptsize, anchor=west] at (0.835,0.342) {Dominated};
            \node[font=\scriptsize, anchor=west] at (0.835,0.307) {Selected};
    
            \node[font=\scriptsize] at (0.165,0.135) {\(\min J\)};
            \node[font=\scriptsize] at (0.340,0.140) {\(\min \tilde{J}\)};
            \node[font=\scriptsize] at (0.505,0.135) {Knee};
            \node[font=\scriptsize] at (0.680,0.135) {Filler};
            \node[font=\scriptsize] at (0.865,0.135) {\(\min J_g\)};
    
            \node[rotate=90] at (0.043, 0.5)
            {\textit{\textbf{Select}} \(N\) trajectories \& \textit{\textbf{re-optimize}}};
    
            \end{scope}
        \end{tikzpicture}
    
        \caption{\textbf{Overview of PER-DDP.} Trajectories optimized with DDP define prior-guided sampling policies, rollouts expand exploration around them, and Pareto filtering selects alternative candidates for reoptimization.}
        \label{fig:pareto_erddp_overview}
        \vspace{-6mm}
    \end{figure}
    
Purely stochastic optimizers provide such exploration by sampling control trajectories broadly as in Cross-Entropy Method (CEM) \cite{kobilarov2012cross}, Model Predictive Path Integral (MPPI) control \cite{williams2017information}, and STOMP \cite{kalakrishnan2011stomp}.
More recently, diffusion-inspired optimizers refine sampled trajectories through annealed noise schedules informed by the dynamics and cost \cite{pan2024modelbased, xue2025full}.
Related population-based optimizers maintain diverse trajectories through Stein variational updates \cite{lambert2021stein, power2024constrained}, explicit multimodal search \cite{osa2020multimodal}, evolution strategies \cite{jankowski2023vpsto}, or massively parallel multi-seed optimization \cite{sundaralingam2023curobo}.

The idea of combining such exploration with DDP dates to guided policy search and its maximum-entropy iLQG variant  \cite{levine2013guided}. 
Sampling-augmented DDP methods later formalized and extended this construction. In particular, Maximum Entropy DDP (ME-DDP) \cite{so2022maximum} casts it as a trajectory optimizer, Generalized Maximum Entropy DDP (GME-DDP) \cite{aoyama2024generalized} broadens the induced policy family, and Stein Variational DDP (SVDDP)~\cite{aoyama2025SVI} uses repulsive interactions to preserve trajectory diversity.
Despite their different mechanisms, these methods share an optimize--explore cycle in which a small set of refined trajectories defines the sampling modes, rollouts are generated around them, and selected ones are promoted for reoptimization.
However, as the same population drives both exploration and optimization, sampling concentrates around a few modes, breadth stays tied to the number of retained trajectories, and heuristics such as elite preservation and mode resampling become necessary.
This limitation sharpens in constrained TO, where ranking rollouts by a single aggregate cost discards trajectories that make meaningful progress in task performance or constraint satisfaction alone, exactly the alternatives that might later mature into high-quality solutions.
 
We address these limitations by introducing \textbf{\underline{P}areto-Optimal \underline{E}ntropy-\underline{R}egularized DDP (PER-DDP)},
a prior-guided population framework that treats sampling modes, rollout evaluations, and retained candidates as independent budgets and selects candidates in the task-constraint objective space rather than through a scalarized cost (Fig. \ref{fig:pareto_erddp_overview}). 
Its sampling policies follow from the free-energy/relative-entropy principle, generalizing maximum-entropy DDP to flexible priors, while Pareto-based selection guarantees a retained trajectory no worse in augmented cost than the DDP solutions.
%
%
%
%
The main contributions of this work are:
    \begin{itemize}
        \item We derive a variational generalization of entropy-regularized DDP from the free-energy/relative-entropy principle, yielding prior-guided policies for bounded-uniform, Gaussian, and multimodal Gaussian priors.
        \item We propose a Pareto-based selection principle for constrained TO that preserves task-constraint alternatives while guaranteeing no degradation in augmented cost relative to DDP at selection steps.
        \item We then present the PER-DDP algorithm which explicitly decouples sampling modes, rollout budget, and retained trajectories, allowing sampling effort to scale independently of the optimization population.
        \item We demonstrate consistent success-rate improvements on bicycle, quadrotor, and Franka Panda systems against state-of-the-art sampling-augmented TO methods, including environments that no baseline solves.
    \end{itemize}

\section{Preliminaries}

    \subsection{Notation}
    \label{sec:notation}

        Scalars, vectors, and matrices are denoted by lowercase, bold lowercase, and bold uppercase letters, respectively, \textit{e.g.}, \(x\in\mathbb{R}\), \(\mathbf{x}\in\mathbb{R}^{n}\), and \(\mathbf{A}\in\mathbb{R}^{m\times n}\). State/control derivatives of $B$ use subscripts, \textit{e.g.}, \(B_x,B_u\) for gradients and \(B_{xx},B_{ux}\) for Hessians. We use \((\cdot)^\top\) for transpose, \(\mathbf{A}\succ0\) for positive definiteness, \(\|\cdot\|\) for the Euclidean norm, and \(\|\mathbf{v}\|_{\mathbf{A}}^2=\mathbf{v}^\top\mathbf{A}\mathbf{v}\) for the weighted norm. We write \(\mathcal{N}(\boldsymbol{\mu},\boldsymbol{\Sigma})\) for a Gaussian with mean \(\boldsymbol{\mu}\) and covariance \(\boldsymbol{\Sigma}\), \(\mathbb{E}_q[\cdot]\) for expectation under \(q\), \(\mathcal{H}(q)\) for differential entropy, and \(D_{\mathrm{KL}}(q\|p)\) for Kullback--Leibler divergence. Finally, \(\operatorname{tr}(\cdot)\), \(\det(\cdot)\), and \(\log\det(\cdot)\) respectively denote trace, determinant, and log-determinant.

    \subsection{Problem Statement}
    \label{sec:problem_statement}

        We consider finite-horizon TO for nonlinear discrete-time systems with dynamics \(\mathbf{x}_{t+1}=F(\mathbf{x}_t,\mathbf{u}_t)\), \(t=0,\ldots,T-1\), fixed \(\mathbf{x}_0\), and \(\mathbf{x}_t\in\mathbb{R}^{n_x}\), \(\mathbf{u}_t\in\mathbb{R}^{n_u}\). Let \(X\triangleq\{\mathbf{x}_0,\ldots,\mathbf{x}_T\}\) and \(U\triangleq\!\{\mathbf{u}_0,\ldots,\mathbf{u}_{T-1}\}\) denote state and control sequences. About a nominal trajectory \((\bar X,\bar U)\), define \(\delta\mathbf{x}_t\triangleq\mathbf{x}_t-\bar{\mathbf{x}}_t\) and \(\delta\mathbf{u}_t\triangleq\mathbf{u}_t-\bar{\mathbf{u}}_t\). The induced trajectory by \(U\) is evaluated by
        \vspace{-2pt}
        \begin{equation}
            J(X,U)
            =
            \sum_{t=0}^{T-1}\ell(\mathbf{x}_t,\mathbf{u}_t)
            +
            \phi(\mathbf{x}_T),
            \label{eq:trajectory_cost}
        \end{equation}
        and the nominal TO problem is then formulated as follows
        \begin{equation}
        \begin{aligned}
            & \min_{U}\quad J(X,U)\\
            \mathrm{s.t.}\quad
            &\mathbf{x}_{t+1}=F(\mathbf{x}_t,\mathbf{u}_t),
            \quad t=0,\ldots,T-1 .
        \end{aligned}
        \label{eq:nominal_ocp}
        \end{equation}
        Constraints \(g_j(\mathbf{x}_t,\mathbf{u}_t)\leq0\), can be addressed with penalties \(\ell_g\) and \(\phi_g\), giving \(J_g\) as in \eqref{eq:trajectory_cost}. With \(\tilde{\ell}=\ell+\ell_g\) and \(\tilde{\phi}=\phi+\phi_g\), the augmented cost \(\tilde{J}=J+J_g\) yields \eqref{eq:nominal_ocp} with $\tilde{J}$ in place of $J$. Hard feasibility may be enforced separately.

    \subsection{Differential Dynamic Programming}
    \label{sec:ddp}
    
        Developed by Jacobson and Mayne \cite{JacobsonMayne1970DDP}, DDP is a second-order single-shooting method for TO with local quadratic convergence. 
        Its Gauss--Newton variant is commonly referred to as iterative Linear Quadratic Regulator \cite{todorov2005generalized}. DDP realizes dynamic programming locally: Bellman's principle of optimality \cite{bellman1966dynamic}, decomposes \eqref{eq:nominal_ocp} into remaining-horizon subproblems. Their optimal tail cost defines the value function \(V_t(\mathbf{x}_t)\), satisfying the Bellman recursion
            \begin{equation}
            V_t(\mathbf{x}_t)
            =
            \min_{\mathbf{u}_t}
            \left[
            \ell(\mathbf{x}_t,\mathbf{u}_t)
            +
            V_{t+1}\!\left(F(\mathbf{x}_t,\mathbf{u}_t)\right)
            \right],
            \label{eq:Bellman_recursion}
            \end{equation}
        with terminal condition \(V_T(\mathbf{x}_T)=\phi(\mathbf{x}_T)\). Moreover, define the $Q$-function such that \(V_t(\mathbf{x}_t)=\min_{\mathbf{u}_t} Q_t(\mathbf{x}_t,\mathbf{u}_t)\). 
            \begin{equation}
            Q_t(\mathbf{x}_t,\mathbf{u}_t)
            \triangleq
            \ell(\mathbf{x}_t,\mathbf{u}_t)
            +
            V_{t+1}\!\left(F(\mathbf{x}_t,\mathbf{u}_t)\right),
            \label{eq:Q_function}
            \end{equation}
        DDP alternates between a backward pass that computes local control gains and a forward pass that updates the nominal trajectory. Working backward from \(t=T-1\) to \(0\), it quadratically approximates \(Q_t\) around \((\bar{\mathbf{x}}_t,\bar{\mathbf{u}}_t)\),
            \begin{align}
            Q_t(\delta\mathbf{x}_t,\delta\mathbf{u}_t)
            \approx{}& \bar Q_t
            +Q_{x,t}^\top\delta\mathbf{x}_t
            +Q_{u,t}^\top\delta\mathbf{u}_t
            +\frac{1}{2}\delta\mathbf{x}_t^\top Q_{xx,t}\delta\mathbf{x}_t 
            \nonumber
            \\
            &+\delta\mathbf{u}_t^\top Q_{ux,t}\delta\mathbf{x}_t
            +\frac{1}{2}\delta\mathbf{u}_t^\top Q_{uu,t}\delta\mathbf{u}_t .
            \label{eq:Q_exp_general}
            \end{align}
        where derivatives are evaluated at \((\bar{\mathbf{x}}_t,\bar{\mathbf{u}}_t)\). For \(Q_{\mathbf{uu},t}\!\succ\!\!0\), minimizing over \(\delta\mathbf{u}_t\) gives the optimal control gains.
            \begin{equation}
            \begin{aligned}
            \delta\mathbf{u}_t^\star &= \mathbf{k}_t+\mathbf{K}_t\delta\mathbf{x}_t,\\
            \begin{bmatrix}\mathbf{k}_t & \mathbf{K}_t\end{bmatrix}
            &=-Q_{\mathbf{uu},t}^{-1}
            \begin{bmatrix}Q_{\mathbf{u},t} & Q_{\mathbf{ux},t}\end{bmatrix}.
            \end{aligned}
            \label{eq:DDP_gains}
            \end{equation}
        Substituting this control law into \eqref{eq:Q_exp_general} and comparing against \eqref{eq:Bellman_recursion} gives a quadratic approximation of \(V_t\) with coefficients
        \vspace{-1mm}
            \begin{subequations}
            \label{eq:DDP_value_updates}
            \begin{align}
            V_t^{\mathrm{DDP}}
            &=
            \bar Q_t
            -
            \frac{1}{2}
            Q_{\mathbf{u},t}^{\top}
            Q_{\mathbf{uu},t}^{-1}
            Q_{\mathbf{u},t},\\
            V_{\mathbf{x},t}^{\mathrm{DDP}}
            &=
            Q_{\mathbf{x},t}
            -
            Q_{\mathbf{xu},t}
            Q_{\mathbf{uu},t}^{-1}
            Q_{\mathbf{u},t},\\
            V_{\mathbf{xx},t}^{\mathrm{DDP}}
            &=
            Q_{\mathbf{xx},t}
            -
            Q_{\mathbf{xu},t}
            Q_{\mathbf{uu},t}^{-1}
            Q_{\mathbf{ux},t}.
            \end{align}
            \end{subequations}
        The forward pass rolls out the dynamics in~\eqref{eq:nominal_ocp} using the computed gains \(\{\mathbf{k}_t,\mathbf{K}_t\}\) to generate a new trajectory,
            \begin{equation}
            \mathbf{u}_t^{\star}
            =
            \bar{\mathbf{u}}_t+\alpha\mathbf{k}_t
            +\mathbf{K}_t(\mathbf{x}_t^{\star}-\bar{\mathbf{x}}_t),
            \quad
            \mathbf{x}_{t+1}^{\star}=F(\mathbf{x}_t^{\star},\mathbf{u}_t^{\star}),
            \label{eq:ddp_forward_rollout}
            \end{equation}
        with \(\mathbf{x}_0^{\star}=\bar{\mathbf{x}}_0\) and \(\alpha\in(0,1]\) chosen by line search. The accepted \((X^{\star},U^{\star})\) replaces the nominal trajectory \((\bar X, \bar U)\), and both passes repeat iteratively until convergence/termination.

    \subsection{Maximum-Entropy Differential Dynamic Programming}
    \label{sec:me_ddp}
    
        Introduced in \cite{levine2013guided} and formalized in \cite{so2022maximum}, ME-DDP extends DDP by replacing the deterministic minimization in \eqref{eq:Bellman_recursion} with an entropy-regularized policy optimization
            \begin{equation}
            V_t(\mathbf{x}_t)
            =
            \inf_{\pi_t}
            \left\{
            \mathbb{E}_{\pi_t}
            \!\left[
            Q_t(\mathbf{x}_t,\mathbf{u}_t)
            \right]
            -
            \lambda\mathcal{H}(\pi_t)
            \right\},
            \label{eq:soft_bellman}
            \end{equation}
        where \(\mathcal{H}(\pi_t)=-\mathbb{E}_{\pi_t}[\log\pi_t(\mathbf{u}_t|\mathbf{x}_t)]\) and \(\lambda>0\) is the temperature. The optimal policy is the Gibbs distribution \cite{theodorou2012relative}
            \begin{equation}
            \pi_t^\star(\mathbf{u}_t|\mathbf{x}_t)
            =
            Z_t(\mathbf{x}_t)^{-1}
            \exp\!\left(-Q_t(\mathbf{x}_t,\mathbf{u}_t)/\lambda\right),
            \label{eq:me_gibbs_policy}
            \end{equation}
        where \(Z_t(\mathbf{x}_t)=\int\exp\!\left(-Q_t(\mathbf{x}_t,\mathbf{u}_t)/\lambda\right)d\mathbf{u}_t\) is the partition function. The corresponding value function is the negatively-scaled log-partition, \(V_t(\mathbf{x}_t)=-\lambda\log Z_t(\mathbf{x}_t)\). Using the local quadratic model \eqref{eq:Q_exp_general} leads to the following Gaussian policy
            \begin{equation}
            \pi_t^\star(\delta\mathbf{u}_t|\delta\mathbf{x}_t)
            =
            \mathcal{N}\!\left(
            \mathbf{k}_t+\mathbf{K}_t\delta\mathbf{x}_t,\,
            \lambda Q_{\mathbf{uu},t}^{-1}
            \right).
            \label{eq:me_local_gaussian}
            \end{equation}
        Thus, ME-DDP retains the DDP mean and feedback gains \eqref{eq:DDP_gains}, while introducing local exploration through the covariance \(\lambda Q_{\mathbf{uu},t}^{-1}\). Under this local Gaussian approximation, entropy regularization affects only the scalar value term, 
            \begin{equation}
            V_t^{\mathrm{ME}}
            =
            V_t^{\mathrm{DDP}}
            +
            \frac{\lambda}{2}
            \left[
            \log\det(Q_{\mathbf{uu},t})
            -
            n_u\log(2\pi\lambda)
            \right],
            \label{eq:me_value_correction}
            \end{equation}
        while \(V_{\mathbf{x},t}\) and \(V_{\mathbf{xx},t}\) retain the DDP form \eqref{eq:DDP_value_updates}. Hence, ME-DDP utilizes the standard DDP backward pass, while the forward pass replaces \(\delta\mathbf{u}_t^\star\) in \eqref{eq:ddp_forward_rollout} with samples from \eqref{eq:me_local_gaussian}, typically drawn around the current best-cost trajectory.
    
    \subsection{Relaxed Log-Barrier Penalty Function}
    \label{sec:prelim_barrier}
    
        Constraints \(g_j(\mathbf{x}_t)\leq0\) are handled through the slack \(z_j(\mathbf{x}_t)\triangleq-g_j(\mathbf{x}_t)\). The classical log-barrier \cite{nocedal2006numerical} is singular at \(z\!=\!0\) and undefined for \(z\!<\!0\), making it unsuitable for infeasible samples. We use the relaxed log-barrier~\cite{grandia2019feedback}
            \begin{equation}
            B(z;\delta)
            =
            \begin{cases}
            -\log z, & z\geq \delta,\\[1mm]
            -\log\delta
            +\frac{1}{2}\left[\left(\frac{z-2\delta}{\delta}\right)^2-1\right],
            & z<\delta,
            \end{cases}
            \label{eq:relaxed_log_barrier}
            \end{equation}
        where \(\delta\!>\!0\) is the relaxation threshold. Given \eqref{eq:relaxed_log_barrier}, $\ell_g$ is
            \begin{equation}
            \ell_g(\mathbf{x}_t)=\mu\sum_{j=1}^{N_g}B\!\left(-g_j(\mathbf{x}_t);\delta\right),
            \label{eq:barrier_penalty}
            \end{equation}
        where \(N_g\) counts constraints and \(\mu\!>\!0\) is the barrier weight; \(\phi_g\) is analogous. The relaxed branch remains finite and differentiable, satisfying DDP's second-order nature.
                    
    \subsection{Pareto Dominance and Pareto Front}
    \label{sec:prelim_pareto}
    
        \textit{Pareto dominance} is a standard concept in multi-objective optimization  \cite{fletcher2002nonlinear}. Let \(m\) index a finite set of candidates, each with objective vector \(\mathbf{C}^m=(C_1^m,\ldots,C_P^m)\in\mathbb{R}^P\), with all objectives minimized. Candidate \(a\) Pareto-dominates candidate \(b\), denoted \(\mathbf{C}^a\prec_{\mathrm{P}}\mathbf{C}^b\), if, for \(p,q\in\{1,\ldots,P\}\),
            \begin{equation}
            C_p^a\leq C_p^b\ \ \forall p,
            \qquad
            \exists q:\ C_q^a<C_q^b .
            \label{eq:pareto_dominance}
            \end{equation}
        The \textit{Pareto front} is the set of nondominated candidates,
            \begin{equation}
            \mathcal{F}_{\mathrm{P}}
            \triangleq
            \left\{
            m:
            \nexists n \ \mathrm{s.t.}\
            \mathbf{C}^n\prec_{\mathrm{P}}\mathbf{C}^m
            \right\}.
        \label{eq:pareto_front}
        \end{equation}

\section{Pareto Entropy-Regularized DDP}

    \subsection{Information-Theoretic Entropy Regularization}
    \label{sec:info_theoretic_er}
    
        The maximum-entropy formulation of Section~\ref{sec:me_ddp} induces a stochastic sampling policy that promotes exploration, helps break symmetries, and reduces sensitivity to poor local minima \cite{kappen2005path}. We derive a variational generalization of this sampling construction through the classical free-energy/relative-entropy principle \cite{theodorou2012relative}, which provides the basis for the developments that follow. Let \(p(U)\) and \(q(U)\) denote probability densities over control sequences \(U\), with \(q\ll p\) so that \(D_{\mathrm{KL}}(q\|p)\) is well defined. Given \(J(X(U),U)\) and \(\lambda>0\), the free-energy/relative-entropy inequality is
            \begin{equation}
            \label{eq:free_energy_ineq}
            \begin{gathered}
            -\lambda\log\mathbb{E}_{p}\!\left[\exp\!\left(-\frac{1}{\lambda}J(X(U),U)\right)\right] \\
            \leq \mathbb{E}_{q}\!\left[J(X(U),U)\right]+\lambda D_{\mathrm{KL}}\!\left(q(U)\|p(U)\right).
            \end{gathered}
            \end{equation}
        Equation \eqref{eq:free_energy_ineq} motivates optimizing over a control-sequence distribution \(q\), whose shape balances low-cost trajectories against proximity to a reference density \(p\). The prior \(p\) provides the sampling structure and may encode useful directions suggested by the optimizer structure, system dynamics, or environment, while \(\lambda\) controls the tradeoff between expected cost and deviation from this prior. The optimal policy is obtained by minimizing the right-hand-side of \eqref{eq:free_energy_ineq}
            \begin{equation}
            q^\star\triangleq\arg\min_{q\ll p}\left\{\mathbb{E}_q[J(X(U),U)]+\lambda D_{\mathrm{KL}}(q\|p)\right\}.
            \label{eq:qstar_variational}
            \end{equation}
        The minimizer is the Gibbs distribution
            \begin{equation}
            q^\star(U)=Z^{-1}\exp\!\left(-J(X(U),U)/\lambda\right)p(U),
            \label{eq:qstar_gibbs}
            \end{equation}
        with partition function \(Z\triangleq\int\exp\!\left(\frac{-J(X(U),U}{\lambda}\right)p(U)\,dU\).
        Substituting \eqref{eq:qstar_gibbs} into \eqref{eq:free_energy_ineq} satisfies the equality,
            \begin{equation}
            -\lambda\log Z=\mathbb{E}_{q^\star}[J(X(U),U)]+\lambda D_{\mathrm{KL}}(q^\star\|p).
            \label{eq:free_energy_equality}
            \end{equation}
        This variational form is closely related to information-theoretic stochastic optimal control, \textit{e.g.,} MPPI \cite{williams2017information}, where a reference sampling law is exponentially reweighted by trajectory cost. We adopt the same principle; however, we choose \(p(U)\) to exploit the structure of the underlying optimizer, here DDP, yielding an optimization-aware \(q^\star(U)\).
    
    \subsection{Local DDP Form}
    \label{sec:local_ddp_form}

        The variational principle in \eqref{eq:qstar_variational} is posed over complete control sequences. Using Bellman's principle of optimality, the trajectory cost \(J\) is decomposed through the state-control function \(Q_t\) in \eqref{eq:Q_function}, yielding the stagewise variational update
            \begin{equation}
            q_t^\star
            =
            \arg\min_{q_t\ll p_t}
            \left\{
            \mathbb{E}_{q_t}\!\left[Q_t(\mathbf{x}_t,\mathbf{u}_t)\right]
            +
            \lambda D_{\mathrm{KL}}(q_t\|p_t)
            \right\}.
            \label{eq:local_er_problem}
            \end{equation}
        To connect the stagewise update to DDP, we use the local quadratic model \eqref{eq:Q_exp_general}. Completing the square in \(\delta\mathbf{u}_t\) gives
            \begin{equation}
            Q_t(\delta\mathbf{x}_t,\delta\mathbf{u}_t)
            =
            \frac{1}{2}\|\delta\mathbf{u}_t-\delta\mathbf{u}_t^\star\|_{Q_{\mathbf{uu},t}}^2
            +\xi_t(\delta\mathbf{x}_t),
            \label{eq:local_q_control_form}
            \end{equation}
        where \(\xi_t(\delta\mathbf{x}_t)\) is independent of \(\delta\mathbf{u}_t\), and \(\delta\mathbf{u}_t^\star\) is given by \eqref{eq:DDP_gains}. Using \eqref{eq:local_q_control_form} as the stagewise counterpart of \(J\) in \eqref{eq:qstar_gibbs}, and absorbing \(\xi_t(\delta\mathbf{x}_t)\) into the partition function $Z$, yields
            \begin{equation}
            q_t^\star(\delta\mathbf{u}_t|\delta\mathbf{x}_t)
            \!=\!
            \frac{1}{Z_t}\exp\!\!\left(-\frac{\|\delta\mathbf{u}_t-\delta\mathbf{u}_t^\star\|_{Q_{\mathbf{uu},t}}^2}{2\lambda}\right)
            \!p_t(\delta\mathbf{u}_t)
            \label{eq:local_er_gibbs}
            \end{equation}
        where  all densities are evaluated at fixed \(\delta\mathbf{x}_t\), with this dependence suppressed where convenient for readability, and 
            \begin{equation}
            Z_t
            =
            \int\exp\!\left(-\frac{\|\delta\mathbf{u}_t-\delta\mathbf{u}_t^\star\|_{Q_{\mathbf{uu},t}}^2}{2\lambda}\right)
            p_t(\delta\mathbf{u}_t)\,d\delta\mathbf{u}_t .
            \label{eq:local_partition}
            \end{equation}
         Notably \(\delta\mathbf{u}_t^\star\) centers the local quadratic cost, and \(Q_{\mathbf{uu},t}\) determines its exploration geometry while \(p_t\)  shapes the resulting \(q_t^\star\). We next specialize \eqref{eq:local_er_gibbs} to tractable policies.

    \subsection{Baseline Priors, Policy Forms and Backward Pass}
    \label{sec:er_prior_selection}

    
        \subsubsection{Bounded Uniform Baseline}
        \label{sec:bounded_uniform_baseline}

            A bounded-uniform (BU) baseline provides a non-informative prior over a prescribed bounded region \(\Omega_t\) of local control increments. In \eqref{eq:local_er_problem}, minimizing \(D_{\mathrm{KL}}(q_t\|p_t)\) relative to this baseline drives \(q_t\) toward a flat distribution on \(\Omega_t\), thereby promoting maximum spread, or equivalently, maximum entropy. Uniform references also appear in earlier maximum-entropy formulations \cite{so2022maximum,levine2013guided}, but cannot be normalized on unbounded control spaces. Restricting \(p_t\) to \(\Omega_t\) therefore provides a proper prior for the KL formulation and, through the absolute continuity condition \(q_t\ll p_t\), also confines \(q_t^\star\) to this region.
            Let
                \begin{equation}
                    p_t(\delta\mathbf{u}_t)
                    =
                    |\Omega_t|^{-1}
                    \mathbf{1}_{\Omega_t}(\delta\mathbf{u}_t),
                    \label{eq:uniform_baseline}
                \end{equation}
            where \(\mathbf{1}_{\Omega_t}\) denotes the indicator of \(\Omega_t\). Substituting \eqref{eq:uniform_baseline} into \eqref{eq:local_er_gibbs}, with \(Z_{\Omega,t}\) denoting normalization over \(\Omega_t\), gives
                \begin{equation}
                q_t^\star(\delta\mathbf{u}_t|\delta\mathbf{x}_t)
                =
                Z_{\Omega,t}^{-1}
                \exp\!\left(-\frac{\|\delta\mathbf{u}_t-\delta\mathbf{u}_t^\star\|_{Q_{\mathbf{uu},t}}^2}{2\lambda}\right)
                \mathbf{1}_{\Omega_t}.
                \label{eq:uniform_truncated_policy}
                \end{equation} 
            Hence, the BU baseline retains the ME-DDP Gaussian kernel \eqref{eq:me_local_gaussian} with covariance \(\lambda Q_{\mathbf{uu},t}^{-1}\), truncated to \(\Omega_t\). Since \(D_{\mathrm{KL}}(q_t^\star\|p_t)=-\mathcal{H}(q_t^\star)+\log|\Omega_t|\), \eqref{eq:free_energy_equality} yields the value term
                \begin{equation}
                V_t^{\mathrm{BU}}(\delta\mathbf{x}_t)
                =
                \mathbb{E}_{q_t^\star}\!\left[Q_t(\delta\mathbf{x}_t,\delta\mathbf{u}_t)\right]
                -\lambda\mathcal{H}(q_t^\star)+\lambda\log|\Omega_t|.
                \label{eq:uniform_er_value}
                \end{equation}
            This matches the maximum-entropy form \cite{so2022maximum} up to \(\lambda\log|\Omega_t|\), which leaves local value derivatives unchanged. For value propagation, we assume weak truncation, with \(\delta\mathbf{u}_t^\star\) far from the boundary of \(\Omega_t\) relative to the Gaussian spread. The policy then retains approximately the Gaussian mean \(\mathbb{E}_{q_t^\star}[\delta\mathbf{u}_t]\approx\delta\mathbf{u}_t^\star\) and covariance \(\operatorname{Cov}_{q_t^\star}[\delta\mathbf{u}_t]\approx\lambda Q_{\mathbf{uu},t}^{-1}\), yielding the DDP value derivatives \eqref{eq:DDP_value_updates}. Under this approximation, BU recovers ME-DDP with a proper least-informative prior and a Gaussian policy truncated to \(\Omega_t\).
            

        \subsubsection{Gaussian Baseline}
        \label{sec:gaussian_baseline}
            We next consider a Gaussian (G) baseline, allowing an informative affine prior control law,
                \begin{equation}
                p_t(\delta\mathbf{u}_t|\delta\mathbf{x}_t)
                =
                \mathcal{N}(\boldsymbol{\mu}_{p,t},\boldsymbol{\Sigma}_{p,t}),
                \quad \!\!\!\!
                \boldsymbol{\mu}_{p,t}\triangleq\mathbf{k}_{p,t}+\mathbf{K}_{p,t}\delta\mathbf{x}_t.
                \label{eq:gaussian_baseline}
                \end{equation}
            Substitution into \eqref{eq:local_er_gibbs} yields a Gaussian product and hence
                \begin{equation}
                q_t^\star(\delta\mathbf{u}_t|\delta\mathbf{x}_t)
                =
                \mathcal{N}(\boldsymbol{\mu}_{q,t},\boldsymbol{\Sigma}_{q,t}),
                \label{eq:gaussian_posterior_policy}
                \end{equation}
            with covariance $\boldsymbol{\Sigma}_{q,t}$ and mean $\boldsymbol{\mu}_{q,t}$ given by the relations
                \begin{subequations}
                \label{eq:gaussian_posterior}
                \begin{align}
                \boldsymbol{\Sigma}_{q,t}
                &=
                \left(\boldsymbol{\Sigma}_{p,t}^{-1}
                +\lambda^{-1}Q_{\mathbf{uu},t}\right)^{-1},
                \label{eq:gaussian_posterior_cov}\\
                \boldsymbol{\mu}_{q,t}
                &=
                \boldsymbol{\Sigma}_{q,t}
                \left(\boldsymbol{\Sigma}_{p,t}^{-1}\boldsymbol{\mu}_{p,t}
                +\lambda^{-1}Q_{\mathbf{uu},t}\delta\mathbf{u}_t^\star\right).
                \label{eq:gaussian_posterior_mean}
                \end{align}
                \end{subequations}
            Because \(\boldsymbol{\mu}_{p,t}\) and \(\delta\mathbf{u}_t^\star=\mathbf{k}_t+\mathbf{K}_t\delta\mathbf{x}_t\) are affine in \(\delta\mathbf{x}_t\), the optimal mean is likewise affine, \(\boldsymbol{\mu}_{q,t}=\mathbf{k}_{q,t}+\mathbf{K}_{q,t}\delta\mathbf{x}_t\). 
                \begin{equation}
                \begin{aligned}
                \mathbf{k}_{q,t}
                &=
                \boldsymbol{\Sigma}_{q,t}
                \left(
                \boldsymbol{\Sigma}_{p,t}^{-1}\mathbf{k}_{p,t}
                +\lambda^{-1}Q_{\mathbf{uu},t}\mathbf{k}_t
                \right),\\
                \mathbf{K}_{q,t}
                &=
                \boldsymbol{\Sigma}_{q,t}
                \left(
                \boldsymbol{\Sigma}_{p,t}^{-1}\mathbf{K}_{p,t}
                +\lambda^{-1}Q_{\mathbf{uu},t}\mathbf{K}_t
                \right).
                \end{aligned}
                \end{equation}
            Define the posterior-to-DDP deviations \(\Delta\mathbf{k}_{q,t}\triangleq\mathbf{k}_{q,t}-\mathbf{k}_t\), \(\Delta\mathbf{K}_{q,t}\triangleq\mathbf{K}_{q,t}-\mathbf{K}_t\), and the prior-to-posterior deviations \(\Delta\mathbf{k}_{p,t}\triangleq\mathbf{k}_{p,t}-\mathbf{k}_{q,t}\), \(\Delta\mathbf{K}_{p,t}\triangleq\mathbf{K}_{p,t}-\mathbf{K}_{q,t}\). Then \(\boldsymbol{\mu}_{q,t}-\delta\mathbf{u}_t^\star=\Delta\mathbf{k}_{q,t}+\Delta\mathbf{K}_{q,t}\delta\mathbf{x}_t\) and \(\boldsymbol{\mu}_{p,t}-\boldsymbol{\mu}_{q,t}=\Delta\mathbf{k}_{p,t}+\Delta\mathbf{K}_{p,t}\delta\mathbf{x}_t\). Let \(\Gamma_t\triangleq\operatorname{tr}(\boldsymbol{\Sigma}_{p,t}^{-1}\boldsymbol{\Sigma}_{q,t})-n_u+\log\frac{\det\boldsymbol{\Sigma}_{p,t}}{\det\boldsymbol{\Sigma}_{q,t}}\). Combining the contributions from \(\mathbb{E}_{q_t^\star}[Q_t]\) and \(D_{\mathrm{KL}}(q_t^\star\|p_t)\), and writing \(V_t^{\mathrm{G}}=V_t^{\mathrm{DDP}}+\Delta V_t^{\mathrm{G}}\) with analogous derivative corrections, matching terms in \(\delta\mathbf{x}_t\) to the local quadratic expansion of \(V_t^{\mathrm{G}}\) yields the following value relations:
                \begin{subequations}
                \label{eq:gaussian_value_corrections}
                \begin{align}
                \Delta V_t^{\mathrm{G}}
                &=\frac{1}{2}\operatorname{tr}(Q_{\mathbf{uu},t}\boldsymbol{\Sigma}_{q,t})
                +\frac{1}{2}\|\Delta\mathbf{k}_{q,t}\|_{Q_{\mathbf{uu},t}}^2 \notag\\
                &\quad+\frac{\lambda}{2}\left(\Gamma_t+\|\Delta\mathbf{k}_{p,t}\|_{\boldsymbol{\Sigma}_{p,t}^{-1}}^2\right),\\
                \Delta V_{\mathbf{x},t}^{\mathrm{G}}
                &=\Delta\mathbf{K}_{q,t}^{\top}Q_{\mathbf{uu},t}\Delta\mathbf{k}_{q,t}
                +\lambda\Delta\mathbf{K}_{p,t}^{\top}\boldsymbol{\Sigma}_{p,t}^{-1}\Delta\mathbf{k}_{p,t},\\
                \Delta V_{\mathbf{xx},t}^{\mathrm{G}}
                &=\Delta\mathbf{K}_{q,t}^{\top}Q_{\mathbf{uu},t}\Delta\mathbf{K}_{q,t}
                +\lambda\Delta\mathbf{K}_{p,t}^{\top}\boldsymbol{\Sigma}_{p,t}^{-1}\Delta\mathbf{K}_{p,t}
                \end{align}
                \end{subequations}
            Unlike the BU case, a Gaussian baseline with state-dependent prior feedback can modify \(V_t^{\mathrm{G}}\), \(V_{\mathbf{x},t}^{\mathrm{G}}\), and \(V_{\mathbf{xx},t}^{\mathrm{G}}\) through differences between the prior, posterior, and DDP gains in \eqref{eq:gaussian_value_corrections}. A useful special case chooses the prior mean at the DDP center, \(\boldsymbol{\mu}_{p,t}=\delta\mathbf{u}_t^\star\). From \eqref{eq:gaussian_posterior_mean}, this gives \(\boldsymbol{\mu}_{q,t}=\delta\mathbf{u}_t^\star\), so the mean-dependent corrections in \eqref{eq:gaussian_value_corrections} vanish, while the KL term still regularizes the policy covariance through \eqref{eq:gaussian_posterior_cov}. Thus, the Gaussian baseline can control exploration variance around the DDP update without shifting its local mean, and its covariance may be annealed across optimization iterations 

        \subsubsection{Multimodal Gaussian Prior}
        \label{sec:mog_baseline}
        
            A single Gaussian prior represents one local basin. To represent multiple basins, let \(r\in\{1,\ldots,R\}\) index local expansions and consider the multimodal Gaussian (MG) prior given by the finite mixture
            \vspace{-1mm}
                \begin{equation}
                p_t=\sum_{r=1}^{R}\pi_r p_t^{(r)},
                \qquad
                \sum_{r=1}^{R}\pi_r=1,
                \label{eq:mm_total_prior}
                \end{equation}
            where \(p_t^{(r)}=\mathcal{N}(\boldsymbol{\mu}_{p,t}^{(r)},\boldsymbol{\Sigma}_{p,t}^{(r)})\) and \(\boldsymbol{\mu}_{p,t}^{(r)}=\mathbf{k}_{p,t}^{(r)}+\mathbf{K}_{p,t}^{(r)}\delta\mathbf{x}_t^{(r)}\). Superscript \((r)\) denotes quantities relative to mode \(r\), while \(\pi_r\) is its prior weight, with \(\pi_r=1/R\) unless otherwise specified. Substituting \eqref{eq:mm_total_prior} into \eqref{eq:local_er_gibbs} and applying the local quadratic model to each component preserves the mixture structure. Each \(p_t^{(r)}\) yields \(q_t^{\star,(r)}\) from \eqref{eq:gaussian_posterior}, with gains and value corrections from \eqref{eq:gaussian_value_corrections}. It remains to determine the posterior mode weights \(w_t^{(r)}\). Define the local partition factor
            {\setlength{\abovedisplayskip}{5pt}\setlength{\belowdisplayskip}{5pt}
                \begin{equation}
                Z_t^{(r)}
                \triangleq
                \int p_t^{(r)}
                \exp\!\left(-Q_t^{(r)}/\lambda\right)
                d\delta\mathbf{u}_t^{(r)} .
                \label{eq:mm_mode_evidence}
                \end{equation}}
            From \eqref{eq:local_er_gibbs} and \eqref{eq:mm_mode_evidence}, each mode satisfies \(p_t^{(r)}e^{-Q_t^{(r)}/\lambda}=Z_t^{(r)}q_t^{\star,(r)}\), giving \(\bar q_t=\sum_{r=1}^{R}\pi_r Z_t^{(r)}q_t^{\star,(r)}\), the unnormalized posterior. Since each \(q_t^{\star,(r)}\) integrates to one, the unnormalized mode masses are \(\pi_r Z_t^{(r)}\). Normalizing gives the weights
            {\setlength{\abovedisplayskip}{5pt}\setlength{\belowdisplayskip}{5pt}
                \begin{equation}
                \begin{aligned}
                q_t^\star
                &=\sum_{r=1}^{R}w_t^{(r)}q_t^{\star,(r)},\\
                w_t^{(r)}
                &=
                \frac{\pi_r Z_t^{(r)}}
                {\sum_{\ell=1}^{R}\pi_\ell Z_t^{(\ell)}}
                =
                \frac{\pi_r e^{-V_t^{\mathrm{G},(r)}/\lambda}}
                {\sum_{\ell=1}^{R}\pi_\ell e^{-V_t^{\mathrm{G},(\ell)}/\lambda}},
                \end{aligned}
                \label{eq:mm_policy}
                \end{equation}}
            where the second equality follows from \eqref{eq:free_energy_equality}, since \(Z_t^{(r)}=\exp[-V_t^{\mathrm{G},(r)}/\lambda]\). Each mode retains its own \(V_t^{\mathrm{G},(r)}\), \(V_{\mathbf{x},t}^{\mathrm{G},(r)}\), and \(V_{\mathbf{xx},t}^{\mathrm{G},(r)}\) from \eqref{eq:gaussian_value_corrections}. The scalar \(V_t^{\mathrm{G},(r)}\) determines \(Z_t^{(r)}\), and hence \(w_t^{(r)}\), while the local derivatives remain associated with their respective \(\delta\mathbf{x}_t^{(r)}\) coordinates. Thus, $q_t^\star$ defines a sampling policy over the local Gaussian posteriors, while the backward-pass approximations remain mode-specific. Sampling selects \(r\) according to \(w_t^{(r)}\) and draws from \(q_t^{\star,(r)}\).
            
    \subsection{Handling Constraints}
    \label{sec:handling_constraints}

        \subsubsection{Control Constraints}
        \label{sec:control_constraints}

            We restrict attention to box control constraints
            \(\Omega_t=\{\delta\mathbf{u}_t:\mathbf{u}_{\min}\leq\bar{\mathbf{u}}_t+\delta\mathbf{u}_t\leq\mathbf{u}_{\max}\}\). We avoid cost-penalty methods for these bounds, since they distort \(Q_{\mathbf{uu},t}\), whose inverse is used directly in the sampling covariance. For fixed \(\delta\mathbf{x}_t\), define \(\mathbf{h}_t(\delta\mathbf{x}_t)\triangleq Q_{\mathbf{u},t}+Q_{\mathbf{ux},t}\delta\mathbf{x}_t\). The constrained local mean $\hat{\delta\mathbf{u}}_t$ is obtained from solving
                \begin{equation}
                \label{eq:control_parametric_qp}
                \hat{\delta\mathbf{u}}_t(\delta\mathbf{x}_t)
                =\arg\min_{\delta\mathbf{u}_t\in\Omega_t}\!
                \bigl\{\tfrac12\|\delta\mathbf{u}_t\|_{Q_{\mathbf{uu},t}}^2
                +\mathbf{h}_t(\delta\mathbf{x}_t)^\top\delta\mathbf{u}_t\bigr\}.
                \end{equation}
            One may solve this state-parametric QP while preserving the affine form \(\hat{\delta\mathbf{u}}_t(\delta\mathbf{x}_t)\approx\mathbf{k}_t+\mathbf{K}_t\delta\mathbf{x}_t\), \textit{e.g.,} with an interior-point treatment, but this relaxes the KKT complementarity optimality condition \cite{aoyama2024second}. We instead use a projection-based update \cite{tassa2014control,bertsekas1982projected}, where we define the local active set required for projecting by solving \eqref{eq:control_parametric_qp} at \(\delta\mathbf{x}_t=\mathbf{0}\), yielding
                \begin{equation}
                    \mathbf{k}_t
                    =
                    \arg\min_{\delta\mathbf{u}_t\in\Omega_t}
                    \left\{
                    \frac{1}{2}
                    \|\delta\mathbf{u}_t\|_{Q_{\mathbf{uu},t}}^2
                    +
                    Q_{\mathbf{u},t}^\top
                    \delta\mathbf{u}_t
                    \right\}.
                    \label{eq:control_box_qp}
                \end{equation}
            Afterwards, the feedback gains \(\mathbf{K}_t\) are reconstructed from the free and active sets using projected-gradient information \cite{tassa2014control,bertsekas1982projected}. The constrained policy \(\hat{\delta\mathbf{u}}_t(\delta\mathbf{x}_t)\approx\mathbf{k}_t+\mathbf{K}_t\delta\mathbf{x}_t\) is affine for a fixed active set and piecewise affine in \(\delta\mathbf{x}_t\). 

        \subsubsection{State Constraints}
        \label{sec:state_constraints}

            State constraints may be handled in DDP through penalties, augmented Lagrangian terms, projections, or barrier methods~\cite{aoyama2024second}. We use the augmented costs of Section~\ref{sec:problem_statement} with the relaxed log barrier of Section~\ref{sec:prelim_barrier}.

    \subsection{Population-Based Policy Evaluation and Mode Selection}
    \label{sec:population_policy_evaluation}

    

        Existing sampling-augmented DDP variants typically use an optimize--sample--reoptimize cycle in which the same finite trajectory set serves as both local solutions and sampling modes. After several DDP iterations, the best-cost trajectory is retained as an elite, while the remaining modes generate only a few rollouts, often one per mode, that are promoted directly for reoptimization~\cite{so2022maximum,aoyama2024generalized,aoyama2025SVI}. Hence, the number of sampling modes, rollouts, and retained trajectories remain coupled, limiting rollout evaluation before reoptimization.

        We decouple these roles. Let \(R\), \(M\), and \(N\) denote the numbers of sampling modes, total stochastic rollouts, and retained DDP candidates, respectively. Each iteration draws \(M\) rollouts from \(R\) modes for evaluation in objective space, from which \(N\) candidates are selected for reoptimization.
            
        \subsubsection{Sampling Modes \(R\)}
        \label{sec:sampling_modes}

            At iteration \(d\), let \(\mathcal{R}_d=\{(\bar{X}_d^{\,r},\bar{U}_d^{\,r})\}_{r=1}^{R}\) denote the modes defining the sampling policies. Each mode induces a local policy \(q_t^{\star,(r)}\), as shown in Section~III.C, around its nominal trajectory. Thus, \(R\) controls the number of local trajectory regions used for sampling.

        \subsubsection{Rollout Samples \(M\)}
        \label{sec:rollout_samples}
    
            The \(M\) rollout samples are distributed across the \(R\) modes to probe their local sampling policies and collected in \(\mathcal{S}_d=\bigcup_{r=1}^{R}\mathcal{S}_d^{\,r}\), with \(|\mathcal{S}_d|=M\). Each \(\mathcal{S}_d^{\,r}\) includes the zero-noise DDP rollout of mode \(r\).

        \subsubsection{Pareto-Optimal Candidate Selection \(N\)}
        \label{sec:pareto_candidate_selection}

            Candidate selection uses realized rollout costs. For each \(s\in\mathcal{S}_d\), let \(J^s\) denote the task cost in \eqref{eq:trajectory_cost}, \(J_g^s\) the constraint-penalty cost from Section~\ref{sec:problem_statement}, \(\tilde{J}^s\triangleq J^s+J_g^s\) the total cost, and \(\mathbf{C}^s\triangleq(J^s,J_g^s)\) its objective vector. We Pareto-filter \(\{\mathbf{C}^s\}_{s\in\mathcal{S}_d}\) using \eqref{eq:pareto_dominance}--\eqref{eq:pareto_front}. From the resulting Pareto front \(\mathcal{F}_{\mathrm{P}}(\mathcal{S}_d)\), we select \(N\) candidates for reoptimization according to the finite-population rule in Section~\ref{sec:candidate_selection}. Notably, Pareto filtering preserves the minimum total-cost rollout.

            \textbf{Proposition 1.} If \(s^\star\in\arg\min_{s\in\mathcal{S}_d}\tilde{J}^s\), then \(s^\star\in\mathcal{F}_{\mathrm{P}}(\mathcal{S}_d)\).
            
            \textit{Proof.} Suppose \(s^\star\notin\mathcal{F}_{\mathrm{P}}(\mathcal{S}_d)\). Then there exists \(n\in\mathcal{S}_d\) such that \(\mathbf{C}^n\prec_{\mathrm{P}}\mathbf{C}^{s^\star}\), implying \(J^n\leq J^{s^\star}\) and \(J_g^n\leq J_g^{s^\star}\), with at least one strict inequality. Therefore, \(\tilde{J}^n<\tilde{J}^{s^\star}\), contradicting \(s^\star\in\arg\min_{s\in\mathcal{S}_d}\tilde{J}^s\). \(\square\)
            
            If the zero-noise rollout \(s_{\mathrm{DDP}}^{(r)}\) of each mode is included in \(\mathcal{S}_d\), Proposition~1 gives
            \(\min_{s\in\mathcal{F}_{\mathrm{P}}(\mathcal{S}_d)}\tilde{J}^s
            =\min_{s\in\mathcal{S}_d}\tilde{J}^s
            \leq\min_{r}\tilde{J}^{s_{\mathrm{DDP}}^{(r)}}\).
            Thus, Pareto filtering will not worsen the best total rollout cost relative to deterministic DDP solutions. 


    \subsection{Forward Pass}
    \label{sec:er_forward_pass}
    
        The backward pass yields \(\{\mathbf{k}_{q,t}^{\,r},\mathbf{K}_{q,t}^{\,r},\boldsymbol{\Sigma}_{q,t}^{\,r}\}_{t=0}^{T-1}\) per mode. We illustrate the forward pass in two steps to clarify the noise injection process \eqref{eq:sampled_forward_rollout_equiv}. First, with line-search parameter \(\alpha\), \eqref{eq:ddp_forward_rollout} yields the optimized deterministic (mean) trajectory:
        {\setlength{\abovedisplayskip}{5pt}\setlength{\belowdisplayskip}{5pt}
            \begin{equation}         \mathbf{u}_t^{\star,r}=\bar{\mathbf{u}}_t^{\,r}+\alpha\mathbf{k}_{q,t}^{\,r}+\mathbf{K}_{q,t}^{\,r}\delta\mathbf{x}_t^{\star,r},
            \quad\!\!\!
            \mathbf{x}_{t+1}^{\star,r}=F(\mathbf{x}_t^{\star,r},\mathbf{u}_t^{\star,r})
            \label{eq:mean_forward_rollout}
            \end{equation}
       where \(\delta\mathbf{x}_t^{\star,r}\triangleq\mathbf{x}_t^{\star,r}-\bar{\mathbf{x}}_t^{\,r}\). Second, stochastic rollouts are generated around this mean by drawing \(\boldsymbol{\epsilon}_t^{\,r,m}\sim\mathcal{N}(\mathbf{0},\boldsymbol{\Sigma}_{q,t}^{\,r})\):
            \begin{equation}
            \mathbf{u}_t^{r,m}
            \!\!=
            \mathbf{u}_t^{\star,r}
            \!\!\!+\alpha\boldsymbol{\epsilon}_t^{r,m}
            \!\!+\mathbf{K}_{q,t}^{r}\delta\mathbf{x}_t^{r,m}, \quad\!\!\!\mathbf{x}_{t+1}^{r,m}\!=\!F(\mathbf{x}_t^{r,m}\!\!,\mathbf{u}_t^{r,m})
            \label{eq:sampled_forward_rollout}
            \end{equation}
        with \(\delta\mathbf{x}_t^{\,r,m}\!\triangleq\!\mathbf{x}_t^{\,r,m}\!\!-\!\mathbf{x}_t^{\star,r}\). Substituting \eqref{eq:mean_forward_rollout} into \eqref{eq:sampled_forward_rollout} gives
            \begin{equation}
            \mathbf{u}_t^{\,r,m}
            =\bar{\mathbf{u}}_t^{\,r}
            +\alpha(\mathbf{k}_{q,t}^{\,r}+\boldsymbol{\epsilon}_t^{\,r,m})
            +\mathbf{K}_{q,t}^{\,r}(\mathbf{x}_t^{\,r,m}-\bar{\mathbf{x}}_t^{\,r}).
            \label{eq:sampled_forward_rollout_equiv}
            \end{equation}}
        Notice that \(\alpha\in(0,1]\) scales both the feedforward step and perturbation, while \(\mathbf{K}_{q,t}^{r}\) provides feedback stabilization.
        \vspace{-2mm}

\vspace{3mm}
\section{Algorithm}
\label{sec:algorithm_design}


    We summarize the proposed PER-DDP in Algorithm~\ref{alg:algorithm}. Let \(\mathcal{T}_d=\{\mathcal{T}_d^n\}_{n=1}^{N}\), with \(\mathcal{T}_d^n=(\bar{X}_d^{\,n},\bar{U}_d^{\,n})\), denote the \(N\) retained candidates at DDP iteration \(d\in\{0,\ldots,D-1\}\), and \(\widehat{\mathcal{T}}_d=\{\widehat{\mathcal{T}}_d^n\}_{n=1}^{N}\) their mean rollouts from \eqref{eq:mean_forward_rollout}. Selected local modes are \(\mathcal{R}_d=\{\widehat{\mathcal{T}}_d^r\}_{r=1}^{R}\subseteq\widehat{\mathcal{T}}_d\). For brevity, \(\{\cdot\}_t\) denotes a sequence over \(t=0,\ldots,T-1\). Table~\ref{tab:algorithm_operators} lists all operators, (our new ones bolded) and Fig~\ref{fig:pareto_erddp_overview} illustrates the framework. 
        \begin{algorithm}[H]
        \caption{PER-DDP}
        \label{alg:algorithm}
        \footnotesize
        \begin{algorithmic}[1]
        
        \Statex \textbf{Input:}
        \(\mathbf{x}_0,\bar{U}_0,T,D,(N,R,M),\lambda,\Omega_t\), DDP-Prior-Barrier parameters
        
        \Statex \vspace{-4pt}
        \Statex \textit{Cold initialization}
        \State \((\bar{X}_0,\bar{U}_0)
        \gets \textsc{Rollout}(\mathbf{x}_0,\bar{U}_0)\)
        \State \(\mathcal{T}_0
        \gets
        \{\mathcal{T}_0^n=(\bar{X}_0,\bar{U}_0)\}_{n=1}^{N}\),
        \(\mathcal{F}_0\gets\emptyset\)
        
        \For{\(d=0,\ldots,D-1\)}
        
            \Statex \vspace{-3pt}
            \Statex \textit{Backward pass and mean rollout}
        
            \For{\(n=1,\ldots,N\)} \Comment{Parallel across Candidates}
                \State \(\{\mathbf{k}_{q,t}^{\,n},
                \mathbf{K}_{q,t}^{\,n},
                \boldsymbol{\Sigma}_{q,t}^{\,n}\}_{t}
                \gets
                \textsc{BackwardPass}(\mathcal{T}_d^n)\)
        
                \State \(\widehat{\mathcal{T}}_d^n
                \gets
                \textsc{MeanRollout}
                (\mathcal{T}_d^n,
                \{\mathbf{k}_{q,t}^{\,n},\mathbf{K}_{q,t}^{\,n}\}_{t})\)
        
                \State \(\textsc{Evaluate}(\widehat{\mathcal{T}}_d^n)\)
            \EndFor
        
            \State \(\widehat{\mathcal{T}}_d
            \gets
            \{\widehat{\mathcal{T}}_d^n\}_{n=1}^{N}\)
        
            \Statex \vspace{-3pt}
            \Statex \textit{Sampling-mode selection}
            \State \(\mathcal{R}_d
            \gets
            \textsc{SelectModes}(\widehat{\mathcal{T}}_d,R)\),
            with \(\mathcal{R}_d\subseteq\widehat{\mathcal{T}}_d\)
        
            \Statex \vspace{-3pt}
            \Statex \textit{Sampling and stochastic rollout}
            \State \(\mathcal{S}_d\gets\emptyset\)
        
            \ForAll{\(\widehat{\mathcal{T}}_d^r\in\mathcal{R}_d\)} \Comment{Parallel across Modes}
                \For{\(m=1,\ldots,M\)} \Comment{Parallel across Rollouts}
                    \State \(\{\boldsymbol{\epsilon}_t^{\,r,m}\}_{t}
                    \gets
                    \textsc{SampleNoise}
                    (\{\boldsymbol{\Sigma}_{q,t}^{\,r}\}_{t})\)
        
                    \State \(\mathcal{S}_d^{r,m}
                    \gets
                    \textsc{NoiseRollout}
                    (\widehat{\mathcal{T}}_d^r,
                    \{\mathbf{K}_{q,t}^{\,r},
                    \boldsymbol{\epsilon}_t^{\,r,m}\}_{t})\)
        
                    \State \(\textsc{Evaluate}(\mathcal{S}_d^{r,m})\)
        
                    \State \(\mathcal{S}_d
                    \gets
                    \mathcal{S}_d
                    \cup
                    \{\mathcal{S}_d^{r,m}\}\)
                \EndFor
            \EndFor
        
            \Statex \vspace{-3pt}
            \Statex \textit{Pareto pruning and candidate selection}
            \State \(\mathcal{F}_{d+1}
            \gets
            \textsc{Pareto}
            (\mathcal{F}_d\cup\widehat{\mathcal{T}}_d\cup\mathcal{S}_d)\)
        
            \State \(\mathcal{T}_{d+1}
            \gets
            \textsc{SelectCandidates}(\mathcal{F}_{d+1},N)\)
        
        \EndFor
        
        \State \Return \(\mathcal{F}_D\) and
        \(\displaystyle
        \arg\min_{(X,U)\in\mathcal{F}_D}
        \tilde J(X,U)\)

        \end{algorithmic}
        \end{algorithm}
        \vspace{-4mm}
        
        \begin{table}[H]
        \centering
        \setlength{\tabcolsep}{-1pt}
        \renewcommand{\arraystretch}{0.93}
        \setlength{\abovecaptionskip}{1pt}
        \caption{Operators used in Algorithm~\ref{alg:algorithm}.}
        \label{tab:algorithm_operators}
        \vspace{-0.2em}
        \begin{tabular}{@{}p{0.29\linewidth}p{0.66\linewidth}@{}}
        \hline
        \rule{0pt}{2.5ex}\textbf{Operator} & \textbf{Meaning} \\[0.2ex]
        \hline
        \(\textsc{Rollout}\) &
        Integrates the dynamics from \(\mathbf{x}_0\) under \(\bar{U}_0\) \eqref{eq:ddp_forward_rollout}. \\[0.3mm]
        
        \(\textsc{Evaluate}\) &
        Computes \(J\) from \eqref{eq:trajectory_cost}, \(J_g\),
        \(\tilde J\), and \(\mathbf{C}=(J,J_g)\). \\[0.3mm]
        
        \(\textsc{\textbf{Pareto}}\) &
        Returns the nondominated subset using
        \eqref{eq:pareto_dominance}--\eqref{eq:pareto_front}. \\[0.3mm]
        
        \(\textsc{BackwardPass}\) &
        Computes \(\{\mathbf{k}_{q,t}^{\,n},\mathbf{K}_{q,t}^{\,n},
        \boldsymbol{\Sigma}_{q,t}^{\,n}\}_{t}\) from
        \eqref{eq:Q_exp_general}, \eqref{eq:local_er_gibbs}, and the prior in
        Section~\ref{sec:er_prior_selection}. \\[0.3mm]
        
        \(\textsc{MeanRollout}\) &
        Applies the deterministic rollout in \eqref{eq:mean_forward_rollout}. \\[0.3mm]
        
        \(\textsc{SampleNoise}\) &
        Draws \(\boldsymbol{\epsilon}_t^{\,r,m}\) using
        \(\boldsymbol{\Sigma}_{q,t}^{\,r}\). \\[0.3mm]
        
        \(\textsc{NoiseRollout}\) &
        Applies the stochastic rollout in \eqref{eq:sampled_forward_rollout}. \\[0.3mm]
        
        \(\textsc{SelectModes}\) &
        For unimodal policies, selects the mode with \(\arg\min \tilde J\); for the multimodal policy, samples \(R\) modes using \eqref{eq:mm_policy} and a categorical distribution. \\[0.3mm]
                
        \begin{tabular}[t]{@{}l@{}}
            \textsc{\textbf{Select}}\\
            \textsc{\textbf{Candidates}}
        \end{tabular} &
        Returns the \(N\) Pareto-front trajectories retained for the next DDP
        iteration; See Section~\ref{sec:candidate_selection}. \\
        \hline
        \end{tabular}
        \vspace{-2mm}
        \end{table}

    We further describe several implementation choices that support the practical use of the proposed algorithm.
    \vspace{1pt}
        \subsubsection{Candidate Selection Operation}
        \label{sec:candidate_selection}


            \textsc{SelectCandidates} selects \(N\) trajectories as anchors and fillers. Anchors preserve principal search modes across DDP iterations. Those whose zero-noise rollouts remain on \(\mathcal{F}_D\) are retained, then supplemented by \(\arg\min\tilde J\), \(\arg\min J\), and the knee of normalized \(\mathcal{F}_D\), maximizing perpendicular distance from the line joining its endpoints. We prefer the knee over the potentially conservative \(\arg\min J_g\). Fillers complete the selection near anchors to explore promising regions, using Euclidean control distance \(d_U(\tau_i,\tau_j)=(\sum_t\|\mathbf{u}_{i,t}-\mathbf{u}_{j,t}\|^2)^{1/2}\) or covariance weighted distance \(d_\Sigma(\tau_i,\tau_j)=(\sum_t\Delta\mathbf{u}_t^\top\boldsymbol{\Sigma}_{q,t}^{-1}\Delta\mathbf{u}_t)^{1/2}\), with threshold \(\gamma_M\in[4,7]\). Duplicates are removed throughout.
                
        \subsubsection{Regularization}
        \label{sec:backward_pass_regularization}
        
            We regularize \(Q_{\mathbf{uu},t}\) before inversion as \(Q_{\mathbf{uu},t}^{\mathrm{reg}}=Q_{\mathbf{uu},t}+\eta_{\mathrm{reg},t}\mathbf{I}\), where \(\eta_{\mathrm{reg},t}\) is the least nonnegative shift, computed from the smallest eigenvalue, such that \(Q_{\mathbf{uu},t}^{\mathrm{reg}}\succ0\); when active, \(Q_{\mathbf{uu},t}^{\mathrm{reg}}\) replaces \(Q_{\mathbf{uu},t}\).
                
        \subsubsection{\(\mathcal{F}_D\) Capacity}
        \label{sec:front_capacity}
        
            For efficiency, we cap \(|\mathcal{F}_D|\) at \(F_{\max}\). When exceeded, trajectories least recently selected by \(\textsc{SelectCandidates}\) are iteratively removed from \(\mathcal{F}_D\).
        
        \subsubsection{Annealing}
        \label{sec:annealing}
        
            For each local mode \(r\), we use a reduced-variance Gaussian prior with Hessian-aligned precision, \(\boldsymbol{\Sigma}_{p,t}^{-1,r,d}=c_d^{\,r}Q_{\mathbf{uu},t}^{\mathrm{reg},r}\), yielding \(\boldsymbol{\Sigma}_{q,t}^{\,r,d}=(c_d^{\,r}+\lambda^{-1})^{-1}(Q_{\mathbf{uu},t}^{\mathrm{reg},r})^{-1}\). As in~\cite{xue2025full}, \(c_d^{\,r}\) is exponentially annealed across DDP iterations, \textit{e.g.,} \(c_d^{\,r}=c_0^{\,r}\exp(-(D-d)/\beta)\).
        
        \subsubsection{Sampling Frequency \& Rejection Sampling}
        \label{sec:rejection_sampling}
        
            Unlike prior variants with sparse sampling, we sample every iteration, though the frequency is configurable. For box-controls, we approximate the truncated policy over \(\Omega_t\) by rejection sampling with a fixed retry limit and clipping fallback.
        
        \subsubsection{General Remarks}
        \label{sec:general_remarks}
        
        The choices of \(p_t\), \(c_d^{\,r}\), and \(\textsc{SelectCandidates}\) are not restrictive; alternatives may be substituted in Algorithm~\ref{alg:algorithm} according to task requirements.

    
\begin{figure*}[!b]
    \vspace{-3mm}
    \centering
    \setlength{\abovecaptionskip}{0pt}

    \begin{tikzpicture}
        \node[anchor=south west, inner sep=0] (img) at (0,0) {%
            \includegraphics[
                width=\textwidth,
                trim=0 2.5mm 0 16mm,
                clip
            ]{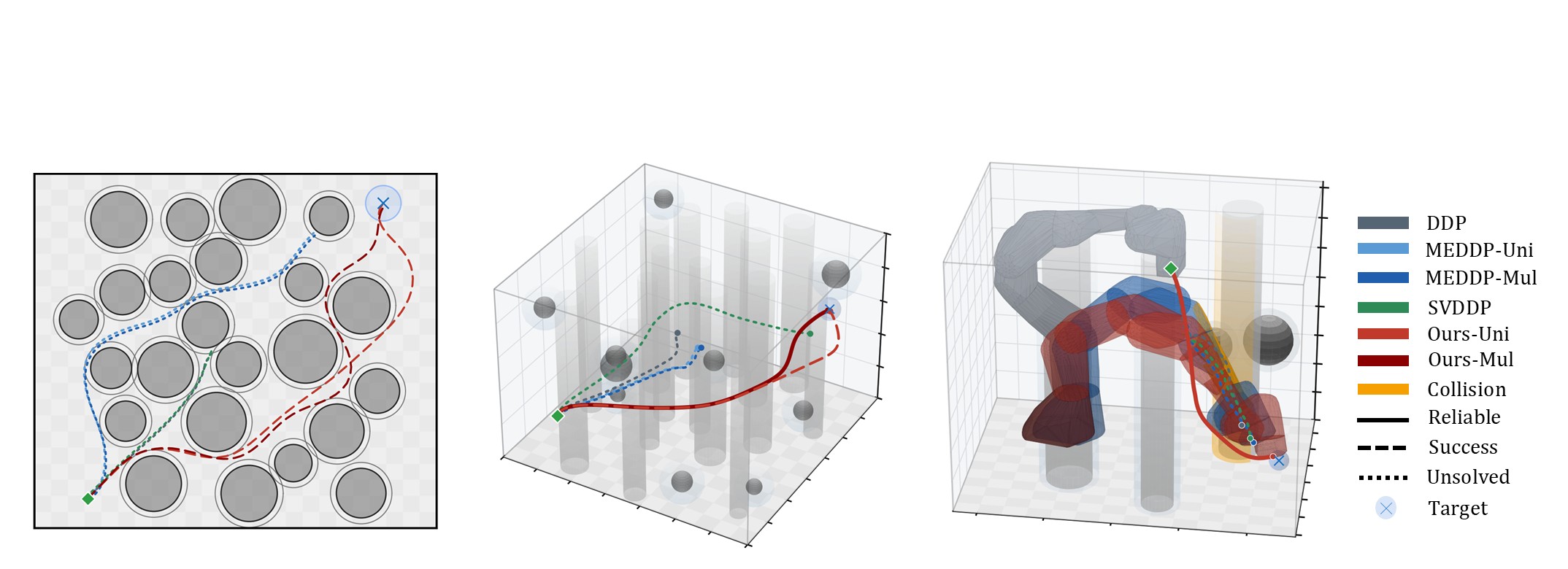}%
        };

        \begin{scope}[
            x={(img.south east)},
            y={(img.north west)}
        ]
            \node[anchor=north, font=\small]
                at (0.150,1.05) {(A) Bicycle};

            \node[anchor=north, font=\small]
                at (0.445,1.05) {(B) Quadrotor};

            \node[anchor=north, font=\small]
                at (0.720,1.05) {(C) Franka};
        \end{scope}
    \end{tikzpicture}

    \caption{\textbf{Representative constrained TO results for (A) BC, (B) QD, and (C) FR}. Only the proposed variants recover feasible solutions across all three environments; the baselines fail to reach the target in (A)--(B) or collide with obstacles in (C), shown in amber. Cylinder clearances omitted for readability.}
    \label{fig:simulation_results}
\end{figure*}


\section{Simulations}
\label{sec:simulations}

    \subsection{Experimental Setup}
    \label{sec:simulation_setup}
    
    The proposed framework combines prior-guided stochastic exploration and Pareto-based population selection to exploit multiple local solutions in non-convex TO. We evaluate it on constrained bicycle (BC), quadrotor (QD), and Franka Panda (FR) problems. The BC follows kinematic bicycle dynamics~\cite{rajamani2012vehicle}, with state \(\mathbf{x}=[p_x,p_y,\theta,v,\psi]^\top\!\!\in\mathbb{R}^{5}\) (position, heading, velocity, steering angle) and control \(\mathbf{u}=[\dot v,\dot{\psi}]^\top\!\!\in\!\mathbb{R}^{2}\). The QD follows the rigid-body dynamics of~\cite{luukkonen2011modelling}, with state \(\mathbf{x}=[\mathbf{p}^\top,\mathbf{v}^\top,\boldsymbol{\eta}^\top,\boldsymbol{\omega}^\top]^\top\!\!\in\!\mathbb{R}^{12}\) (position, velocity, Euler angles, angular velocity) and rotor-force inputs \(\mathbf{u}\in\mathbb{R}^{4}\). The FR is simulated in MuJoCo/MJX~\cite{todorov2012mujoco}, with state \(\mathbf{x}=[\mathbf{q}^\top,\dot{\mathbf{q}}^\top]^\top\in\mathbb{R}^{14}\) (joint positions and velocities) and joint-torque inputs \(\mathbf{u}\!=\!\boldsymbol{\tau}\!\in\!\mathbb{R}^{7}\). We compare against DDP~\cite{JacobsonMayne1970DDP}, uni- and multimodal ME-DDP~\cite{so2022maximum}, and SVDDP~\cite{aoyama2025SVI}.

            \begin{table*}[!t]
            \centering
            \setlength{\abovecaptionskip}{-2pt}
            \caption{Problem and environment specifications. Here, \(a_n\triangleq a\mathbf{1}_n\), and \(\#\) denotes obstacle count.}
            \label{tab:simulation_setup}
        
            \vspace{-1.2mm}
        
            \scriptsize
            \setlength{\tabcolsep}{2.4pt}
            \renewcommand{\arraystretch}{0.92}
            \setlength{\aboverulesep}{0.25ex}
            \setlength{\belowrulesep}{0.25ex}
        
            \begin{tabular*}{\textwidth}{@{\extracolsep{\fill}}lccccc@{}}
                \toprule
        
                \rule{0pt}{2.3ex}
                \textbf{System}
                & \(\boldsymbol{T,\Delta t\, ,D}\)
                & \textbf{Initial \(\boldsymbol{\rightarrow}\) Target}
                & \textbf{Objective Parameters}
                & \textbf{Bounds}
                & \textbf{Obstacles}
                \\[0.3mm]
        
                \midrule
        
                BC
                & \(100,.02,400\)
                & \(
                    \begin{gathered}
                        [0_2,\frac{\pi}{4},0_2]^\top
                        \!\!\rightarrow\!
                        [5_2,\frac{\pi}{4},0_2]^\top
                    \end{gathered}
                  \)
                & \(
                    \begin{gathered}
                        \mathbf Q^\ell=\operatorname{diag}(.5_2,.01,.001,.01)\\[-0.8mm]
                        \mathbf Q^\phi=\operatorname{diag}((2{\times}10^4)_2,(10^3)_2,1)\\[-0.8mm]
                        \mathbf R=\operatorname{diag}(10^{-3},.1)
                    \end{gathered}
                  \)
                & \(
                    \begin{gathered}
                        p_x\in[-0.5,5.5]\\[-0.8mm]
                        p_y\in[-0.5,5.5]\\[-0.8mm]
                        |\dot v|\leq10,\ |\dot\psi|\leq2
                    \end{gathered}
                  \)
                & \(
                    \begin{gathered}
                        \text{circles (M/H/C)}\\[-0.8mm]
                        \!\!\#\!\!:\!\! [15,20)/\!+\!5/\!+\!10\\[-0.8mm]
                        \kappa=.10
                    \end{gathered}
                  \)
                \\
        
                QD
                & \(120,.02\, ,400\)
                & \(
                    \begin{gathered}
                        \mathbf p_0=[0_2,1]^\top
                        \!\!\rightarrow\!
                        \mathbf p^{\mathrm{ref}}=[5_2,3]^\top\\[-0.8mm]
                        \mathbf v^{\mathrm{ref}}
                        =\boldsymbol{\eta}^{\mathrm{ref}}
                        =\boldsymbol{\omega}^{\mathrm{ref}}
                        =\mathbf0
                    \end{gathered}
                  \)
                & \(
                    \begin{gathered}
                        \mathbf Q^\ell=\operatorname{diag}(1_3,.5_3,.1_3,.01_3)\\[-0.8mm]
                        \mathbf Q^\phi=\operatorname{diag}(10^3_3,100_3,50_3,1_3)\\[-0.8mm]
                        \mathbf R=.01\mathbf I_4
                    \end{gathered}
                  \)
                & \(
                    \begin{gathered}
                        p_x\in[-1,6],\ p_y\in[-1,6]\\[-0.8mm]
                        p_z\in[0,5]\\[-0.8mm]
                        |u_i|\leq2.5
                    \end{gathered}
                  \)
                & \(
                    \begin{gathered}
                        \text{cyl./spheres (M/H/C)}\\[-0.8mm]
                        \#:25/30/35\\[-0.8mm]
                        \kappa=.30
                    \end{gathered}
                  \)
                \\
        
                FR
                & \(100,.01\, 100\)
                & \(
                    \begin{gathered}
                        \mathbf q_{\mathrm{home}}
                        \rightarrow
                        \mathbf p_{\mathrm{EE}}^{\mathrm{ref}}\\[-0.8mm]
                        \mathbf p_{\mathrm{EE}}^{\mathrm{ref}}
                        =[.68,-.42,.20]^\top\,\mathrm{m}
                    \end{gathered}
                  \)
                & \(
                    \begin{gathered}
                        \mathbf Q^\ell=\operatorname{diag}(.01_7,.1_7),\quad\!\!\!\!
                        \mathbf Q^\phi=\operatorname{diag}(.01_7,10_7)\\[-0.8mm]
                        \mathbf R=10^{-4}\mathbf I_7,\quad\!\!\!\!\!
                        w_{\mathrm{EE}}^\ell=10,\quad\!\!\!\!
                        w_{\mathrm{EE}}^\phi=1.5{\times}10^4
                    \end{gathered}
                  \)
                & \(
                    \begin{gathered}
                        \mathbf q: \text{ MJX Built-in Bounds \cite{todorov2012mujoco}}\\[-0.8mm]
                        |\tau_{1:4}|\leq87,\ |\tau_{5:7}|\leq12
                    \end{gathered}
                  \)
                & \(
                    \begin{gathered}
                        \text{spheres/pillars}\\[-0.8mm]
                        3\text{ layouts},\quad\!\!\! \kappa=.02
                    \end{gathered}
                  \)
                \\
        
                \bottomrule
            \end{tabular*}

            \vspace{-6mm}
        \end{table*}
    
    Each benchmark is a finite-horizon reaching TO problem with given initial state \(\mathbf{x}_0\) and reference state \(\mathbf{x}^{\mathrm{ref}}\) and costs
        \begin{equation}
        \begin{aligned}
        \ell(\mathbf{x}_t,\mathbf{u}_t)
        &=\tfrac{1}{2}\|\mathbf{x}_t-\mathbf{x}^{\mathrm{ref}}\|_{\mathbf{Q}^{\ell}}^2
        +\tfrac{1}{2}\|\mathbf{u}_t\|_{\mathbf{R}}^2
        +\ell_{\mathrm{EE}}(\mathbf{x}_t),\\
        \phi(\mathbf{x}_T)
        &=\tfrac{1}{2}\|\mathbf{x}_T-\mathbf{x}^{\mathrm{ref}}\|_{\mathbf{Q}^{\phi}}^2
        +\phi_{\mathrm{EE}}(\mathbf{x}_T),
        \end{aligned}
        \label{eq:simulation_costs}
        \end{equation}
    where \(\ell_{\mathrm{EE}}=\phi_{\mathrm{EE}}=0\) for the BC and QD. For FR, \(\mathbf{x}^{\mathrm{ref}}=[\mathbf{q}_{\mathrm{home}}^\top,\mathbf{0}^\top]^\top\), penalizing deviations from the home joint configuration and nonzero joint velocities, while the reaching objective is specified in Cartesian space by the end-effector (EE) target \(\mathbf{p}_{\mathrm{EE}}^{\mathrm{ref}}\). Accordingly, \(\ell_{\mathrm{EE}}(\mathbf{x}_t)=\tfrac12 w_{\mathrm{EE}}^\ell\|\mathrm{FK}(\mathbf{q}_t)-\mathbf{p}_{\mathrm{EE}}^{\mathrm{ref}}\|^2\) and \(\phi_{\mathrm{EE}}(\mathbf{x}_T)=\tfrac12 w_{\mathrm{EE}}^\phi\|\mathrm{FK}(\mathbf{q}_T)-\mathbf{p}_{\mathrm{EE}}^{\mathrm{ref}}\|^2\), where \(w_{\mathrm{EE}}^\ell\), \(w_{\mathrm{EE}}^\phi\) are weights and $ \mathrm{FK}$ the Forward Kinematics. Constraints are handled as in Section~\ref{sec:handling_constraints}. For circular and spherical obstacles, \(g_o(\mathbf{p})=(r_o+\kappa)^2-\|\mathbf{p}-\mathbf{c}_o\|^2\leq0\), where \(\kappa\) is the clearance; cylinders use the same constraint on the position projected onto the plane perpendicular to their axes. We fix \((\mu,\delta)=(10^{-2},10^{-5})\). The BC and QD test sets each comprise 297 environments, with 99 instances per moderate/hard/challenging (M/H/C) obstacle tier, while FR comprises 99 environments with 33 per clutter layout; Table~\ref{tab:simulation_setup} summarizes the benchmark problem specifications.
    
    The proposed method follows Section~\ref{sec:algorithm_design}, setting \(R\!=\!N\) so each retained candidate defines a sampling mode while \(M\) sets the rollout population. We use \((N,M,F_{\max})=(10,100,150)\) for BC/QD and \((5,50,80)\) for FR, with \(\lambda=(5{\times}10^{-3},5{\times}10^{-3},10^{-2})\) and \(\beta=(.85,.8,.9)\), respectively. All baselines share \(N,D\), use individually tuned \(\lambda\), sampling interval \(\Delta_d\), and method-specific parameters, and run \(5/5/3\) times per BC/QD/FR environment without early stopping. We report success rate (SR), reliable success rate (RSR), iterations-to-best (ItB), and unique successes (UNS): SR is the fraction of environments with a feasible run within goal tolerance (\(0.3/0.3/0.03\,\mathrm{m}\)), RSR requires a majority of runs to succeed, ItB is the median iterations to the best solution over reliably solved environments, and UNS counts environments solved by the proposed variant but no baseline.

    \vspace{-0.5mm}
    
    \subsection{Performance Evaluation and Benchmarking}
    \label{sec:simulation_results}

    The purpose of the expanded search is to recover useful trajectories that would otherwise remain unexplored. Across all three systems, PERDDP achieves higher success rates and finds solutions where all baselines fail (Table~\ref{tab:results}). These unique successes extend the set of solvable environments beyond the combined coverage of the baseline methods, supporting the motivation for broader sampling and retaining alternative task--constraint tradeoffs. The accompanying improvement in RSR indicates that the gains persist under repeated runs, while the lowest reported ItB values suggest that principled exploration reaches favorable solutions sooner.

    The representative cases depict how alternative motions can overcome the incomplete or infeasible local solutions common amongst baselines (Fig.~\ref{fig:simulation_results}). In BC and QD (A--B), the proposed variants reach the targets through distinct routes, extending beyond the partial progress of several baselines. In FR (C), competing motions approach the target but intersect the obstacle, whereas PERDDP completes the reach safely. These outcomes highlight that useful alternatives may differ in either task progress or feasibility, motivating a closer look at how candidates develop during optimization.


    Following these candidates across iterations reveals the value of preserving intermediate tradeoffs (Fig.~\ref{fig:demo}). Local refinement stalls near obstacles at \(d=7,184\), but Pareto filtering preserves alternative tradeoffs around which multiple rollouts uncover routes past the obstruction (\(d=8,185\)). Retaining candidates with task or feasibility progress despite higher total cost allows further sampling and reoptimization to build on these escape routes and recover successful solutions (\(d=D\)). The displayed baselines remain incomplete, consistent with exploration restricted by elite preservation and scalarized ranking. Candidates initially considered uncompetitive can prove decisive achieving eventual success.

    \begin{table*}[!t]
        \centering
        \setlength{\abovecaptionskip}{-1pt}
        \caption{Performance metrics across the three benchmark systems.}
        \label{tab:results}
        \vspace{-1.2mm}
    
        \scriptsize
        \setlength{\tabcolsep}{4.0pt}
        \renewcommand{\arraystretch}{0.95}
        \setlength{\aboverulesep}{0.25ex}
        \setlength{\belowrulesep}{0.25ex}
    
        \begin{tabular*}{\textwidth}{@{\extracolsep{\fill}}lcccccccccccc@{}}
            \toprule
            \rule{0pt}{2.5ex}
            & \multicolumn{4}{c}{\textbf{Bicycle}}
            & \multicolumn{4}{c}{\textbf{Quadrotor}}
            & \multicolumn{4}{c}{\textbf{Franka}}
            \\
            \cmidrule(lr){2-5}
            \cmidrule(lr){6-9}
            \cmidrule(lr){10-13}
    
            \textbf{Method}
            & \textbf{SR [\%]} & \textbf{RSR [\%]} & \textbf{UNS} & \textbf{ItB}
            & \textbf{SR [\%]} & \textbf{RSR [\%]} & \textbf{UNS} & \textbf{ItB}
            & \textbf{SR [\%]} & \textbf{RSR [\%]} & \textbf{UNS} & \textbf{ItB}
            \\
            \midrule
    
            DDP
            & 20.2 & -- & 0 & 398
            & 24.9 & -- & 2 & 346
            & 41.4 & -- & 0 & 71
            \\
    
            MEDDP-Uni
            & 47.1 & 37.7 & 0 & 338
            & 51.9 & 40.1 & 0 & 273
            & 51.5 & 49.5 & 0 & 58
            \\
    
            MEDDP-Mul
            & 47.1 & 37.4 & 0 & 340
            & 51.5 & 39.1 & 0 & 269
            & 51.5 & 49.5 & 0 & 58
            \\
    
            SVDDP
            & 53.2 & 42.1 & 6 & 325
            & 54.2 & 43.1 & 3 & 299
            & 49.5 & 49.5 & 0 & 59
            \\
    
            Ours-Uni
            & \textbf{62.6} & \textbf{44.1} & \textbf{16} & 337
            & 58.9 & 47.1 & 18 & 266
            & \textbf{57.6} & \textbf{53.5} & \textbf{3} & \textbf{42}
            \\
    
            Ours-Mul
            & 54.9 & 40.7 & 11 & \textbf{256}
            & \textbf{60.6} & \textbf{47.5} & \textbf{21} & \textbf{265}
            & -- & -- & -- & --
            \\
    
            \bottomrule
        \end{tabular*}
    
        \vspace{-4mm}
    \end{table*}

    \begin{figure*}[t]
        \centering
        \setlength{\abovecaptionskip}{-3pt}
        \includegraphics[
            keepaspectratio,
            trim=0 0mm 0mm 1mm, 
            clip
        ]{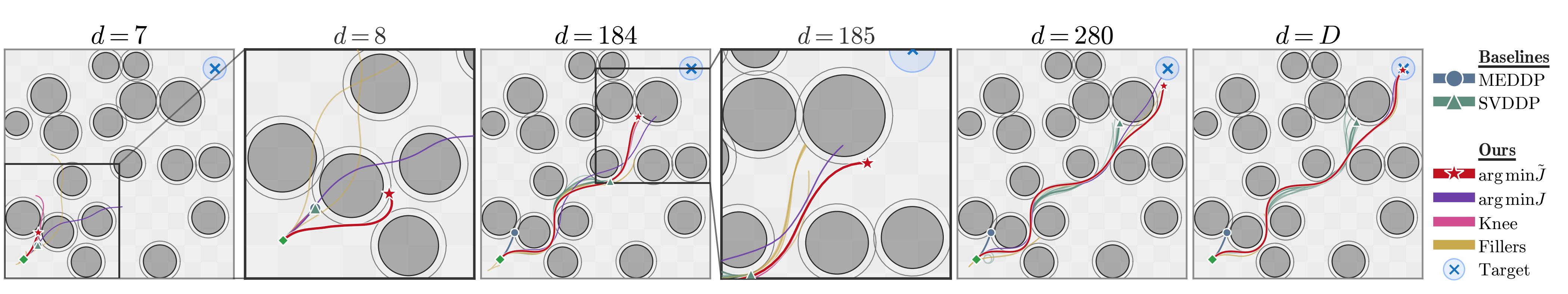}
        \caption{\textbf{The role of Pareto filtering in evading non-convexity.} The sequence tracks PERDDP's retained trajectories alongside MEDDP and SVDDP over optimization iterations, highlighting task--constraint tradeoffs (\(d=8,185\)) that develop into successful trajectories while the baselines remain incomplete.}
        \label{fig:demo}
        \vspace{-3mm}
    \end{figure*}

\section{Conclusion}


We presented PER-DDP, an entropy-regularized population framework for constrained TO that shapes exploration through flexible control priors, preserves task-constraint alternatives via Pareto selection, and decouples rollout evaluation from local optimization. 
On three robotic systems, PER-DDP improved success rates and recovered feasible motions missed by all baselines, validating its balance of local refinement and alternative solution discovery.
Although PER-DDP uses additional rollouts, the number of local DDP solves remains unchanged, and both sampling and local optimization are highly parallelizable. 
While this study focuses on offline tasks in simulation, the parallel structure of PER-DDP makes real-time MPC deployment and hardware validation a natural next step. Further directions include extending PER-DDP to multi-agent settings \cite{saravanos2023distributed} and combining it with learning-augmented sampling-based optimization schemes \cite{saravanos2026learningtooptimizeflows}.


\section*{Acknowledgment}

Dimitrios S. Georgiou gratefully acknowledges financial support from the A. Onassis Foundation Scholarship and thanks Y. Aoyama and M. Jung, for valuable discussions.

\bibliographystyle{IEEEtran}
\bibliography{references}

    
\end{document}